\documentclass[letterpaper, 10 pt, conference]{ieeeconf}  % Comment this line out if you need a4paper

\IEEEoverridecommandlockouts                              % This command is only needed if
\usepackage{graphicx}
\graphicspath{{figs/}}
\usepackage{amsmath}
\usepackage{amssymb}
\usepackage{algorithm}
\usepackage{algpseudocode}
\usepackage{times}
\usepackage{url}
\usepackage{flushend}

\newcommand{\figref}[1]{Fig.~\ref{#1}}
\newcommand{\tabref}[1]{Table~\ref{#1}}

\title{\LARGE \textbf
{
  Self-Body Image Acquisition and Posture Generation with Redundancy using Musculoskeletal Humanoid Shoulder Complex\\for Object Manipulation
}
}

\author{Yuya Koga$^{1}$, Kento Kawaharazuka$^{1}$, Yasunori Toshimitsu$^{1}$, Manabu Nishiura$^{1}$, Yusuke Omura$^{1}$,\\Yuki Asano$^{1}$, Kei Okada$^{1}$, Koji Kawasaki$^{2}$ and Masayuki Inaba$^{1}$% <-this % stops a space
  \thanks{$^{1}$ The authors are with the Department of Mechano-Informatics, Graduate School of Information Science and Technology, The University of Tokyo, 7-3-1 Hongo, Bunkyo-ku, Tokyo, 113-8656, Japan.
    {\texttt\small [koga, kawaharaduka, toshimitsu, nishiura, oomura, asano, k-okada, inaba]@jsk.imi.i.u-tokyo.ac.jp}
  }
  \thanks{$^{2}$ The author is associated with TOYOTA MOTOR CORPORATION.
    {\texttt\small koji\_kawasaki@mail.toyota.co.jp}
  }
}

\begin{document}

\maketitle
\thispagestyle{empty}
\pagestyle{empty}

\begin{abstract}
We proposed a method for learning the actual body image of a musculoskeletal humanoid for posture generation and object manipulation using inverse kinematics with redundancy in the shoulder complex.
The effectiveness of this method was confirmed by realizing automobile steering wheel operation.
The shoulder complex has a scapula that glides over the rib cage and an open spherical joint, and is supported by numerous muscle groups, enabling a wide range of motion.
As a development of the human mimetic shoulder complex, we have increased the muscle redundancy by implementing deep muscles and stabilize the joint drive.
As a posture generation method to utilize the joint redundancy of the shoulder complex, we consider inverse kinematics based on the scapular drive strategy suggested by the scapulohumeral rhythm of the human body.
In order to control a complex robot imitating a human body, it is essential to learn its own body image, but it is difficult to know its own state accurately due to its deformation which is difficult to measure.
To solve this problem, we developed a method to acquire a self-body image that can be updated appropriately by recognizing the hand position relative to an object for the purpose of object manipulation.
We apply the above methods to a full-body musculoskeletal humanoid, Kengoro, and confirm its effectiveness by conducting an experiment to operate a car steering wheel, which requires the appropriate use of both arms.
\end{abstract}

% \begin{IEEEkeywords}
% Biomimetics, Learning from Experience, Bioinspired Robot Learning 
% \end{IEEEkeywords}

\section{INTRODUCTION}
The musculoskeletal humanoid \cite{mizuuchi2004kenta} \cite{mizuuchi2006kotaro} \cite{nakanishi2012kenshiro} \cite{ikemoto2012humanlike} \cite{jantsch2013anthrob} is a humanoid robot that mimicks the joint and drive structures of the human body.
By controlling redundant muscle actuators placed across the passive joints, it is expected to elucidate the mechanisms of the human body and improve the physical capabilities of humanoids.

\begin{figure}[tbh]
 \begin{center}
  \includegraphics[width=1.0\columnwidth]{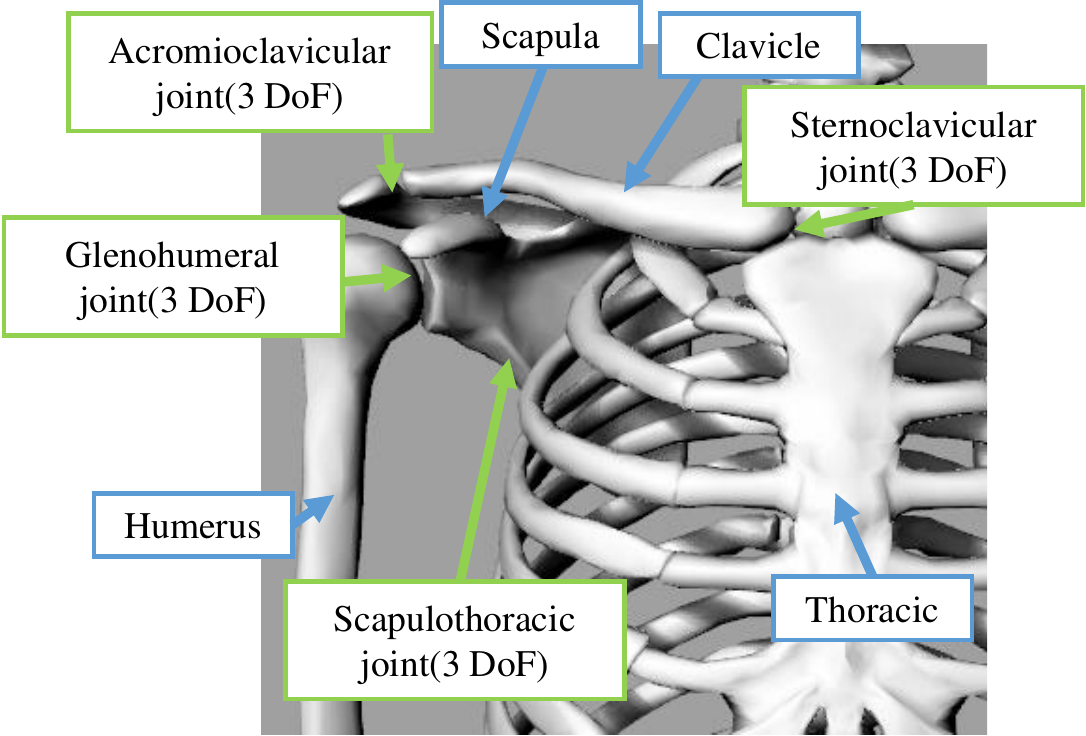}
  \caption{Human Shoulder complex model.}
  \label{figure:human_shoulder_complex}
 \end{center}
\end{figure}

In this study, we focus on the shoulder complex and improve the hardware, and then apply the method to generate posture and learn the self-body image of the actual machine by taking advantage of the joint redundancy.
The shoulder complex shown in Fig. 1 consists of the rib cage, clavicle, scapula, humerus, and four joints connecting them.
It is related to a wide range of movements of the upper limbs and has a unique joint structure in the human body with the scapula and the open spherical joint.
Several attempts have been made to reproduce this open spherical joint, but each has its own issues in terms of stability and structural differences compared to the human body\cite{nakanishi2012kenshiro} \cite{ikemoto2012humanlike}.
Therefore, we are working on hardware design of shoulder complex that focuses on the deep muscles of the human body and aims to achieve stability of open spherical joints.

In addition, the scapula has very high muscle redundancy, and by focusing on this joint group, we expect to be able to devise a method that can be applied to other parts of the body.
In the field of computer graphics, a method for generating natural postures by inverse kinematics based on a database of joint angles of the human body has been proposed\cite{kim2016data}.
Although posture generation methods for redundant structures have been proposed in previous research\cite{maurel2000human}, it is difficult to apply them directly to flexible robots, so we propose a simple method that is inspired by the scapulohumeral rythm\cite{inmanshoulder}.

In control of the musculoskeletal humanoid, it is essential to solve the difficult-to-measure deformations, joint redundancy, and muscle redundancy of robots.
In order to solve the elongation and contraction of robot structures and muscle actuators, a method to obtain the relationship between muscles and joints from actual data using machine learning has been proposed \cite{kawaharazuka2018online}.
On the other hand, it is difficult to apply such a method to musculoskeletal humanoids that have characteristic structures of the human body such as open spherical joints and scapulae.
To solve this problem, we propose a self-body image learning method using object relative hand position recognition for object manipulation.

Finally, we apply these methods to the musculoskeletal humanoid Kengoro \cite{asano2016kengoro} and confirm their effectiveness in a car steering wheel operation experiment.

All human skeleton models in this paper are taken from \cite{humandata}.

\section{Development of human mimetic shoulder complex}
The shoulder complex is characterized by a high degree of joint redundancy, a wide range of motion, and a tendency to become unstable without muscle support.
In this study, we divided the shoulder complex into two parts: the part related to the scapular motion and the drive of the glenohumeral joint, which is called the shoulder joint in a narrow sense, and considered each approach.

\subsection{Stability of scapular drive}
\begin{figure}[tbh]
 \begin{center}
  \includegraphics[angle=270,width=1.0\columnwidth]{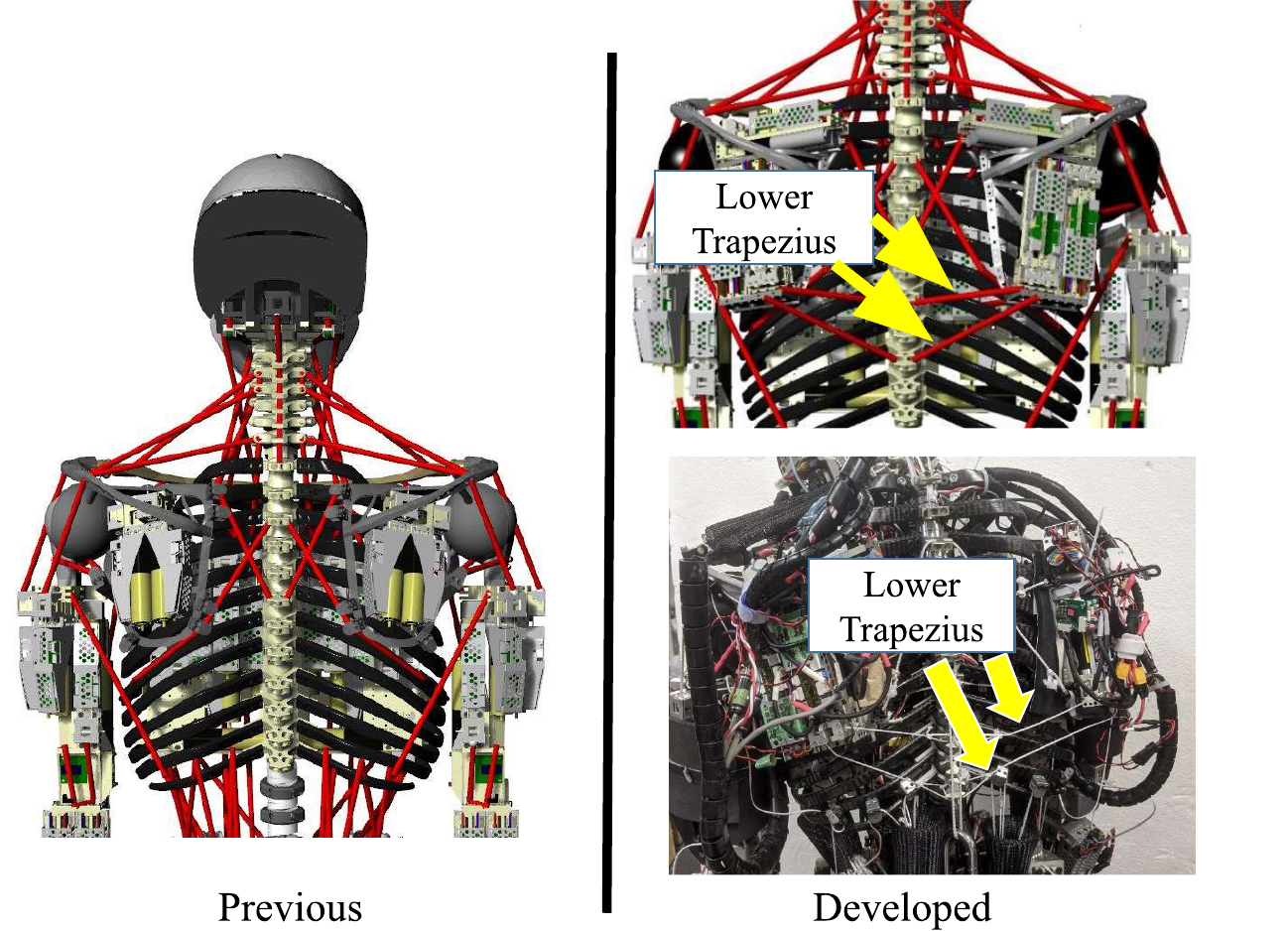}
  \caption{Kengoro trapezius model.}%% in Rhinoceros \cite{rhino}.}
  \label{figure:rhino_trapezius}
 \end{center}
\end{figure}

The scapula moves as if it were sliding on the thorax, and the trapezius muscle, which is a huge areolar muscle, plays a role in pressing and supporting the scapula against the thorax.
In order to replace the planar muscles with muscle actuators, the method of repeatedly folding between multiple members has been studied \cite{nakanishi2012kenshiro}.
However, this method suffers from a delay in response due to folding and uneven loading.
In this study, multiple relatively compact muscle actuators \cite{Kawaharazuka_iros2017} are placed to substitute the function of the trapezius muscle by increasing the direction of the force applied to the scapula.
The specific design is shown in \figref{figure:rhino_trapezius}.
The muscle actuators acting as the lower part of the trapezius muscle are placed on the scapula and connected to the spine respectively.

%% 既存の肩甲骨は、左右それぞれ菱形筋・僧帽筋上部・僧帽筋中部・前鋸筋の4本の筋によって動作していた。
%% 本研究では、新たに僧帽筋下部の二本を追加し、肩甲骨に合計６本の筋がつながる構造とした。

The existing scapula was operated by four muscles: the rhomboid, upper trapezius, middle trapezius, and serratus anterior muscles on the left and right sides, respectively.
In this study, two new muscles, the lower trapezius, were added, making the scapula a structure with a total of six muscles connected to it.

\subsection{Stability of the glenohumeral joint}
\begin{figure}[tbh]
 \begin{center}
  \includegraphics[width=1.0\columnwidth]{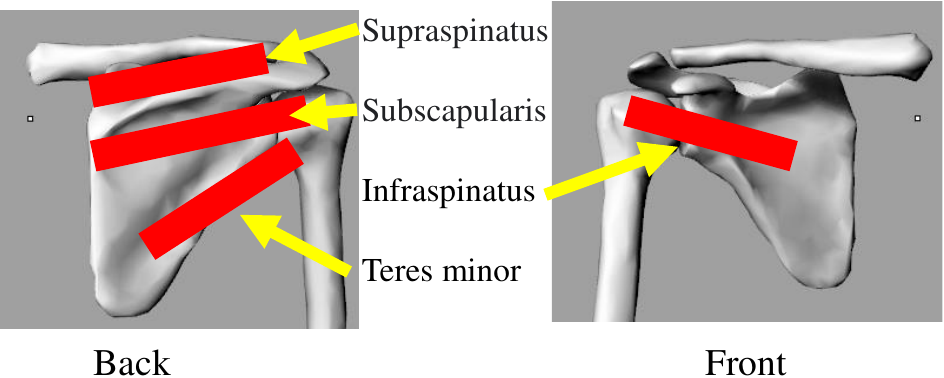}
  \caption{Human rotator cuff model.}
  \label{figure:human_rotator_cuff}
 \end{center}
\end{figure}

The glenohumeral joint is an open spherical joint in which a ball fits into a shallow bowl and has inherent instability.
In order to support this instability, muscle redundancy is high, and shallow and deep muscles are arranged so that they overlap.
The shallow muscles, which are relatively far from the joint and have large moment arms, exert the main force to move the joint.
On the other hand, the deep muscles shown in \figref{figure:human_rotator_cuff} are relatively close to the joint and have a small moment arm, so they contribute to joint stabilization by exerting a force to bring the links closer together instead of joint motion.

\begin{figure}[tbh]
 \begin{center}
  \includegraphics[width=1.0\columnwidth]{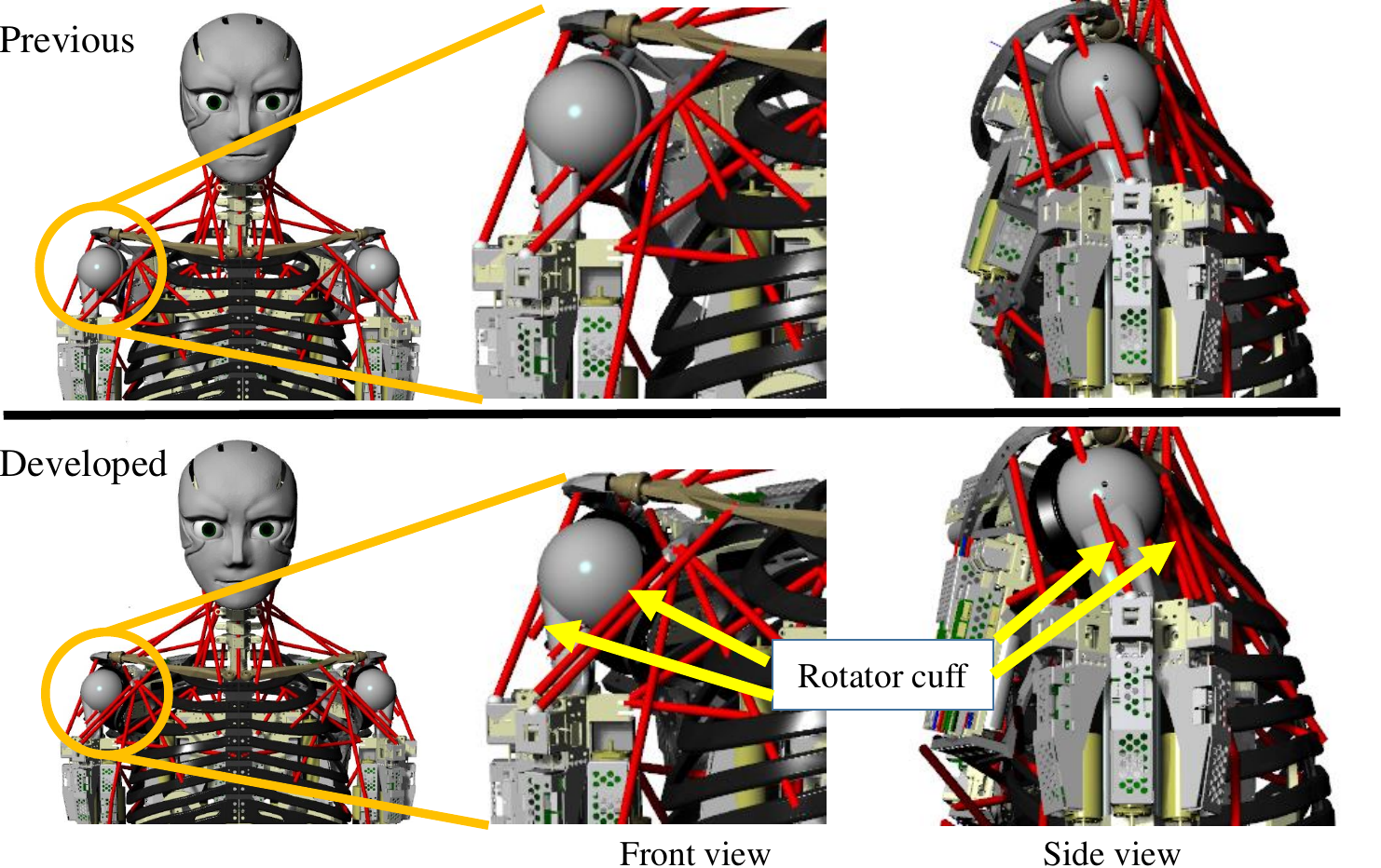}
  \caption{Kengoro Rotator cuff model in Rhinoceros.}
  \label{figure:rhino_rotator_cuff}
 \end{center}
\end{figure}

In musculoskeletal humanoid design, shallow muscles have been emphasized mainly to achieve minimum joint motion due to spatial constraints.
However, the glenohumeral joint, which has a particularly wide range of motion among the shoulder complexes, is an open spherical joint in which a ball is contained in a shallow bowl, and shallow muscles alone can quickly lead to dislocation.
Therefore, a compact module is placed on the scapula to implement the rotator cuff, which is a deep muscle group supporting the glenohumeral joint.
As shown in \figref{figure:rhino_rotator_cuff}, the rotator cuff is placed between the scapula and the ball of the glenohumeral joint to play the role of a deep muscle group in the human body.

%% 既存の肩複合体において、肩甲骨と上腕骨は５本の筋により接続されていた。
%% 本研究の肩複合体では、深層筋二本を追加し合計７本の筋により接続されることとなった。

In the existing shoulder complex, the scapula and humerus were connected by five muscles.
In the shoulder complex of this study, two deep muscles were added, resulting in a total of seven muscle connections.

\subsection{implementation on Kengoro and basic evaluation experiment}
\begin{figure}[tbh]
 \begin{center}
  \includegraphics[width=1.0\columnwidth]{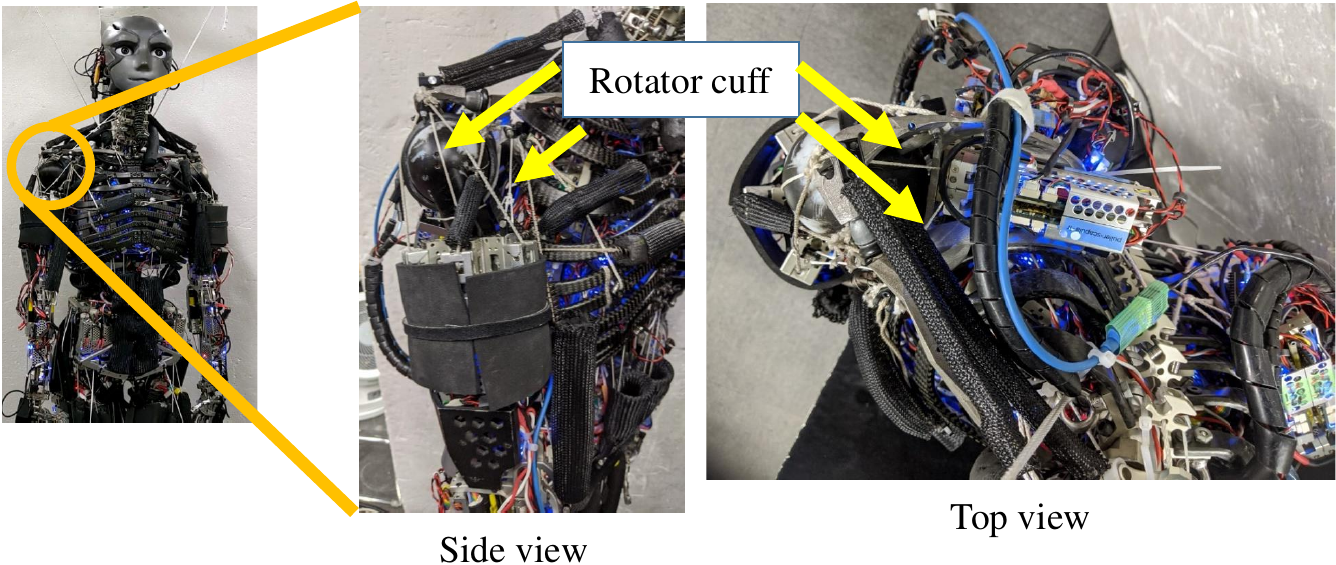}
  \caption{Rotator cuff on Kengoro.}
  \label{figure:kengoro_rotator_cuff}
 \end{center}
\end{figure}

\begin{figure}[tbh]
 \begin{center}
  \includegraphics[angle=270,width=1.0\columnwidth]{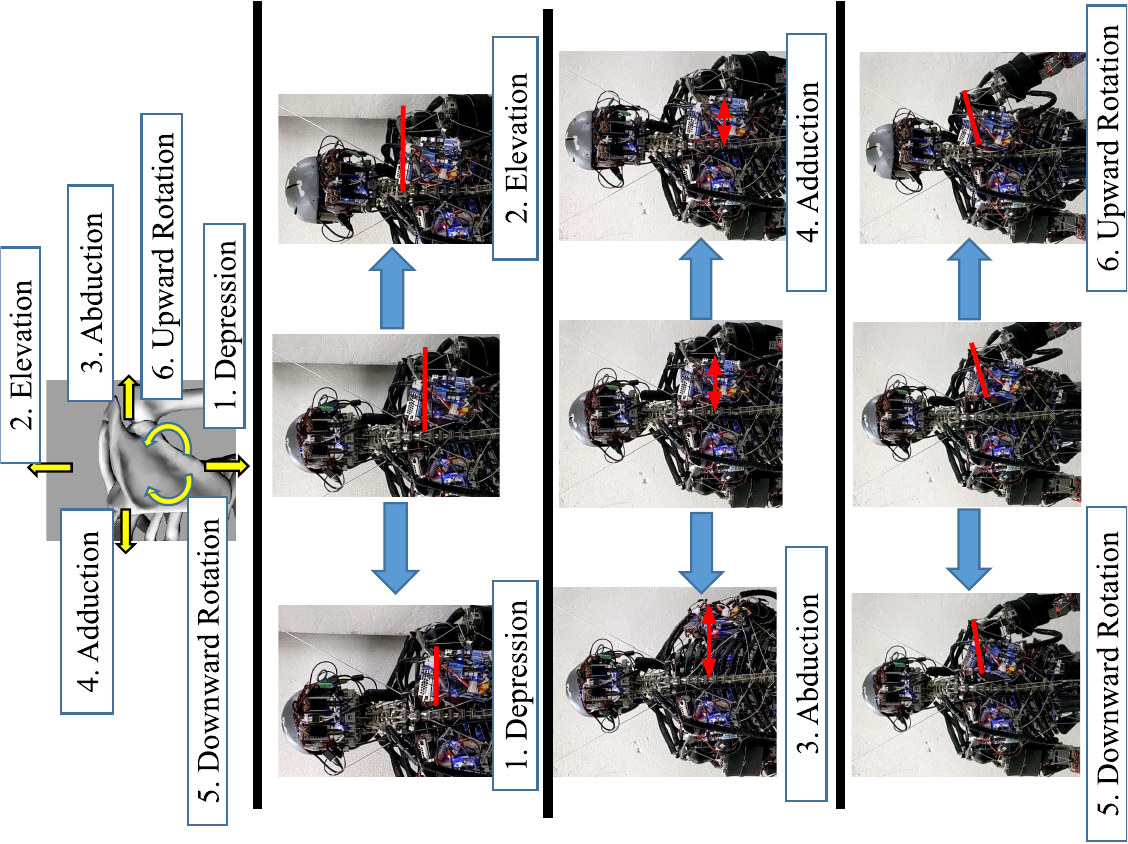}
  %% \caption{Scapula movement experiments\cite{basic_movement_scapula}.}
  \caption{Scapulothoracic joint experiments.}
  \label{figure:basic_exp_scapula}
 \end{center}
\end{figure}

\begin{figure}[tbh]
 \begin{center}
  \includegraphics[angle=270,width=1.0\columnwidth]{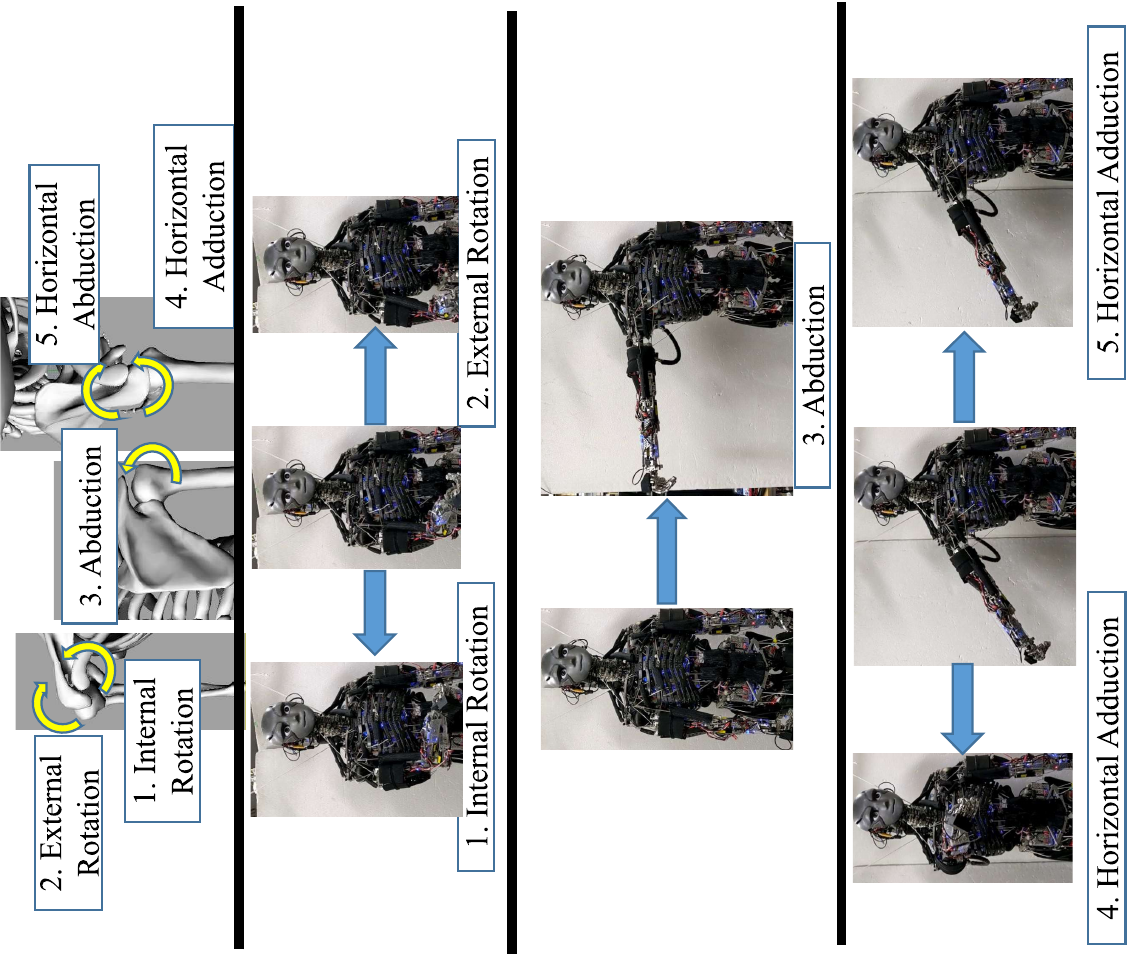}
  %% \caption{Shoulder joint experiments\cite{shoulder_bone_joint_model}.}
  \caption{Glenohumeral joint experiments.}
  \label{figure:basic_exp_shoulder}
 \end{center}
\end{figure}

The shoulder complex designed in this study was mounted on a full-body musculoskeletal humanoid, Kengoro, as shown in \figref{figure:kengoro_rotator_cuff}.
The movements of the scapulothoracic joint were classified into six types as shown in \figref{figure:basic_exp_scapula}, and we confirmed that each of them could be operated on the actual machine.
Five types of basic movements of the glenohumeral joint were performed as shown in \figref{figure:basic_exp_shoulder}, and it was confirmed that the rotator cuff stabilized the joint, especially in the abduction movement.

\section{Posture generation with redundancy}
Posture generation to drive the scapula is considered to be important in upper limb movements using a human-mimetic shoulder complex.
There is a previous study that optimized the inverse kinematics calculation under the condition that the scapula is attached to the rib cage \cite{maurel2000human}.
Such a method requires a detailed model of the robot, and it is difficult to apply it directly to a complex human-mimicking robot in which the thorax itself can be deformed.

\begin{figure}[tbh]
 \begin{center}
  \includegraphics[angle=270,width=1.0\columnwidth]{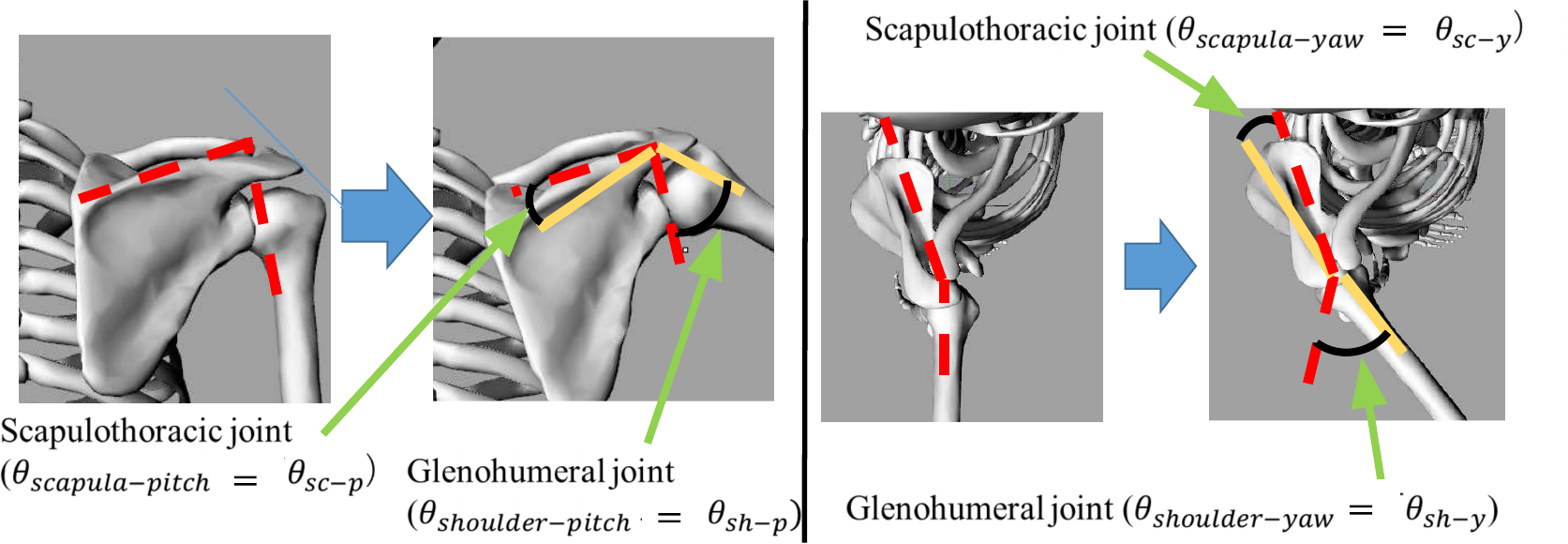}
  \caption{Joint angle definition of shoulder complex flextion/extention.}
  \label{figure:scapulohumeral_rythm}
 \end{center}
\end{figure}

Therefore, in this study, we focus on the scapulohumeral rhythm of the human body and propose a method to generate the posture of the upper limbs including the scapula more simply.
As shown in \figref{figure:scapulohumeral_rythm}, the scapulohumeral rhythm is a mechanism in which a person's scapulothoracic joint angle and scapulohumeral joint angle maintain a certain relationship.
The human body is known to move in such a way that the scapulothoracic joint angle and the glenohumeral joint angle are approximately $1:2$ \cite{inmanshoulder}.
This makes it possible to expand the reach of the hand and reduce the load on the glenohumeral joint, which is prone to dislocation.

The specific procedure of inverse kinematics of the shoulder complex suggested by this scapulohumeral rhythm is as follows.
\begin{itemize}
  \item initial posture \par
  \item Fix the scapula and solve the inverse kinematics using the glenohumeral, elbow and wrist joints \par
  \item Multiply the scapulohumeral joint angle by a constant to determine the scapulothoracic joint angle.
    \begin{equation}
      {\theta}_{scapula\ roll} = A * {\theta}_{shoulder\ roll}
    \end{equation}
    \begin{equation}
      {\theta}_{scapula\ pitch} = A * {\theta}_{shoulder\ pitch}
    \end{equation}
    \begin{equation}
      A = Constant
    \end{equation}
  \item Fix the moved scapula and solve the inverse kinematics using the glenohumeral, elbow and wrist joints again \par
\end{itemize}

\begin{figure}[tbh]
 \begin{center}
  \includegraphics[width=1.0\columnwidth]{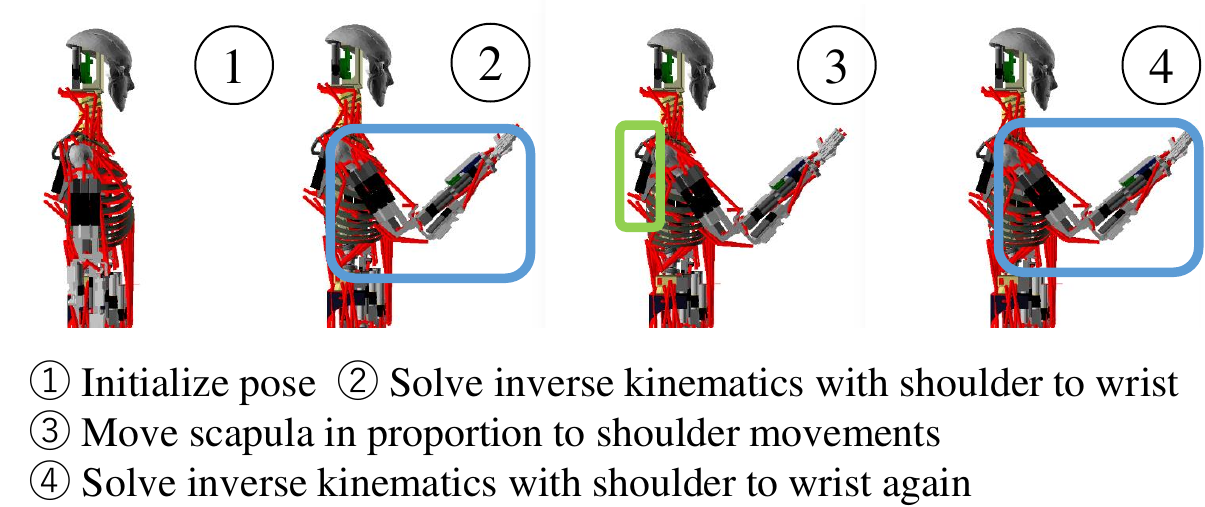}
  \caption{Inversekinematics involving scapula movements.}
  \label{figure:scapula_ik}
 \end{center}
\end{figure}

In this study, $A=1/2.7$ is used as a reference value for the human body.
In other words, the scapulothoracic joint angle and the glenohumeral joint angle move in a ratio of approximately $1:2.7$.
The actual application of this method to Kengoro's geometric model is shown in \figref{figure:scapula_ik}.

The comparison of the joint angles before and after the scapula motion in \figref{figure:scapula_ik} is shown in \tabref{table:scapula_ik}.
The joint angles in \tabref{table:scapula_ik} are the angles shown in \figref{figure:scapulohumeral_rythm} and the elbow joint angle ${\theta}_{elbow}$.
%肩甲骨のわずかな動作により、人体と同様に肩甲上腕関節角度及び肘関節角度の変化が小さくなっている。
%特に肩甲上腕関節は開放型球関節であるため、大きな角度変化による脱臼が発生しやすい。
%従って、本手法によって同じ手先位置を実現する、負荷の低く安定した関節角度を計算できていることがわかる。
The slight movement of the scapula results in a small change in the scapulohumeral joint angle and elbow joint angle, as in the human body.
In particular, since the scapulohumeral joint is an open spherical joint, dislocation due to large angle changes is likely to occur.
Therefore, it can be seen that the present method is able to calculate a stable joint angle with low load that achieves the same hand position.

\begin{table}[htb]
    \begin{center}
        \caption{Comparison of joint angles with and without scapular motion.}
            \begin{tabular}{|c||c|c|c|c|c|} \hline
                & ${\theta}_{sc-r}$ & ${\theta}_{sc-y}$ & ${\theta}_{sh-r}$ & ${\theta}_{sh-y}$ & ${\theta}_{elbow}$ \\ \hline \hline
                %% No scapula \raise0.2ex\hbox{\textcircled{\scriptsize{2}}}[$deg$] & 0.0 & 0.0 & 4.9 & 8.9 & -102.2 \\ \hline
                %% With Scapula\raise0.2ex\hbox{\textcircled{\scriptsize{4}}}[$deg$] & 1.8 & 3.0 & 2.5 & 5.8 & -95.4\\ \hline
                {\scriptsize No scapula[$deg$]} & 0.0 & 0.0 & 4.9 & 8.9 & -102.2 \\ \hline
                {\scriptsize With scapula[$deg$]} & 1.8 & 3.0 & 2.5 & 5.8 & -95.4\\ \hline
            \end{tabular}
            %% \begin{tabular}{|c||c|c|} \hline
            %%     & ${\theta}_{scapula\ roll}$ & ${\theta}_{scapula\ yaw}$ \\ \hline \hline
            %%     No scapula \raise0.2ex\hbox{\textcircled{\scriptsize{2}}}[$deg$] & 0.0 & 0.0 \\ \hline
            %%     With Scapula\raise0.2ex\hbox{\textcircled{\scriptsize{4}}}[$deg$] & 1.8 & 3.0 \\ \hline
            %% \end{tabular}
            %% \begin{tabular}{|c||c|c|c|c|} \hline
            %%     & ${\theta}_{shoulder\ roll}$ & ${\theta}_{shoulder\ yaw}$ & ${\theta}_{elbow}$ \\ \hline \hline
            %%     \raise0.2ex\hbox{\textcircled{\scriptsize{2}}}[$deg$] & 4.9 & 8.9 & -102.2 \\ \hline
            %%     \raise0.2ex\hbox{\textcircled{\scriptsize{4}}}[$deg$] & 2.5 & 5.8 & -95.4 \\ \hline
            %% \end{tabular}
        \label{table:scapula_ik}
    \end{center}
\end{table}
%\begin{table}[htb]
%    \begin{center}
%        \caption{Comparison of joint angles with and without scapular motion.}
%            \begin{tabular}{|c||c|c|} \hline
%                & ${\theta}_{scapula\ roll}$ & ${\theta}_{scapula\ yaw}$ \\ \hline \hline
%                Without scapular motion \raise0.2ex\hbox{\textcircled{\scriptsize{2}}} [degree] & 0.0 & 0.0 \\ \hline
%                With scapular motion\raise0.2ex\hbox{\textcircled{\scriptsize{4}}} [degree] & 1.8 & 3.0 \\ \hline
%            \end{tabular}
%            \begin{tabular}{|c||c|c|c|c|} \hline
%                & ${\theta}_{shoulder\ roll}$ & ${\theta}_{shoulder\ yaw}$ & ${\theta}_{elbow}$ \\ \hline \hline
%                \raise0.2ex\hbox{\textcircled{\scriptsize{2}}} [degree] & 4.9 & 8.9 & -102.2 \\ \hline
%                \raise0.2ex\hbox{\textcircled{\scriptsize{4}}} [degree] & 2.5 & 5.8 & -95.4 \\ \hline
%            \end{tabular}
%            %% \begin{tabular}{|c||c|c|c|c|c|c|} \hline
%            %%     & ${\theta}_{scapula roll}$ & ${\theta}_{scapula yaw}$ & ${\theta}_{shoulder roll}$ & ${\theta}_{shoulder pitch}$ & ${\theta}_{shoulder yaw}$ & ${\theta}_{elbow}$ \\ \hline \hline
%            %%     \raise0.2ex\hbox{\textcircled{\scriptsize{2}}}[$^\circ$] & 0.0 & 0.0 & 4.9 & -33.2 & 8.9 & -102.2 \\ \hline
%            %%     \raise0.2ex\hbox{\textcircled{\scriptsize{4}}}[$^\circ$] & 1.8 & 3.0 & 2.5 & -37.3 & 5.8 & -95.4 \\ \hline
%            %% \end{tabular}
%        \label{table:scapula_ik}
%    \end{center}
%\end{table}

\section{Self-body image acquisition for object manipulation}
In this section, we describe a method that enables a musculoskeletal humanoid robot to learn its own body image for the purpose of object manipulation.

As mentioned in Section I, researches have been conducted to learn the relationship between muscles and joints in musculoskeletal humanoids and to realize their movements.
However, it is difficult to apply this method to the shoulder complex including the scapula, where both joint redundancy and muscle redundancy are very high.
In this study, we devise a method to learn the self-body image by fixing the initial posture and estimating the joint angles for object manipulation.

\subsection{Initial posture definition and estimation of joint angle for object manipulation}
\begin{figure}[tbh]
 \begin{center}
  \includegraphics[width=1.0\columnwidth]{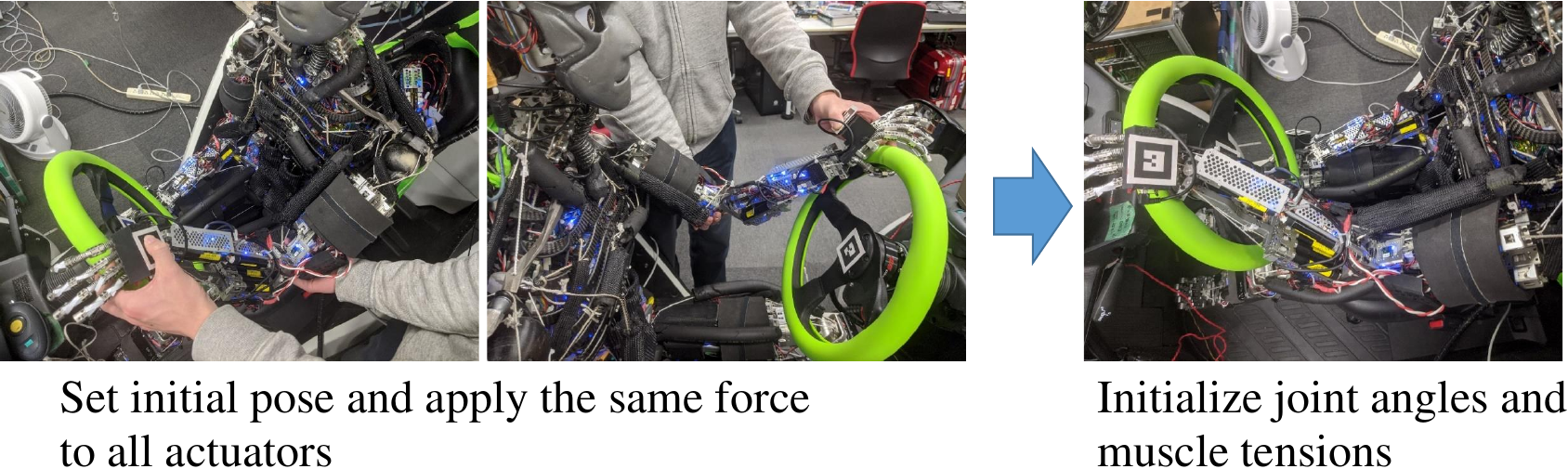}
  \caption{Initial posture determination.}
  \label{figure:kengoro_initialize}
 \end{center}
\end{figure}

In the past, musculoskeletal humanoids have performed all their motions with their arms lowered to the ground in an alert posture, but this posture is not appropriate for the purpose of object manipulation.
Therefore, as shown in \figref{figure:kengoro_initialize}, we decided to use the average posture and the posture that is stable only by adding a certain low tension as the initial posture for all future experiments.
%動作の平均的姿勢をとらせた後、紐のたるみを防ぎ姿勢を固定するためにすべての筋に4kgのちからを加え、安定した状態を姿勢の初期位置とした。
%続いて筋長さを目標とする通常の制御に移行し、以下の操作を開始する。
After having the participants assume the average posture for the operation, 4 kg of force was applied to all the muscles to prevent the muscles from slackening and to fix the posture, and the stable state was set as the initial posture position.
Then, we shifted to the normal control to target the muscle length and started the following operations.

\begin{figure}[tbh]
 \begin{center}
  \includegraphics[width=1.0\columnwidth]{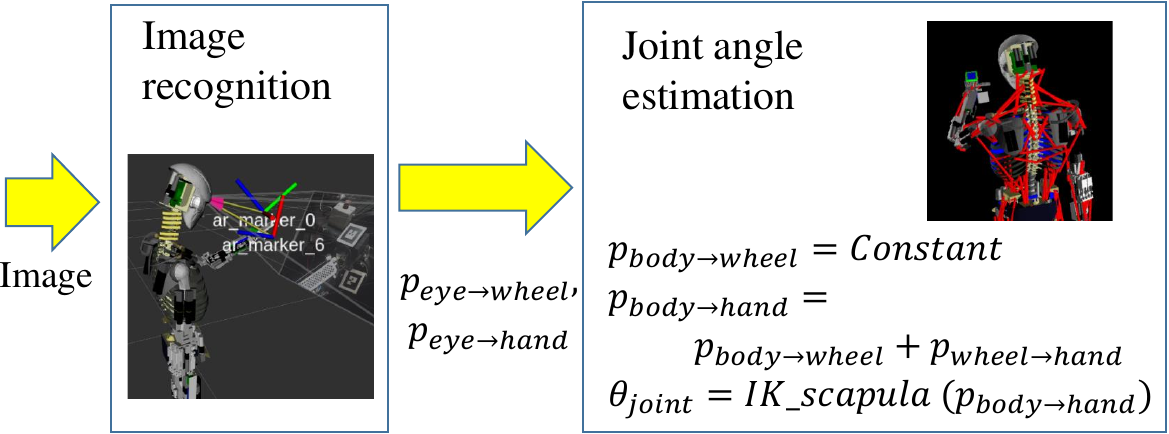}
  \caption{Joint angle estimation method for steering wheel operation.}
  \label{figure:vision_estimator}
 \end{center}
\end{figure}

Next, the calculation method of the joint angle estimation for generating the teacher data in the self-body image acquisition is shown in \figref{figure:vision_estimator}.
From the images acquired by the head-mounted camera, we calculate the three-dimensional coordinates ($p_{eye \rightarrow wheel}, p_{eye \rightarrow hand}$) of the AR markers attached to the hand and the manipulated object with respect to the eyes.
These coordinates include both position and orientation\cite{ar_track_alvar}.
Then, the object position ($p_{body \rightarrow wheel}$) seen from the robot body is known, and the hand position ($p_{body \rightarrow hand}$) seen from the robot body is obtained by adding the hand position relative to the object.
Finally, the inverse kinematics method described in the previous section is applied based on this hand position to calculate the estimated joint angle.

It is important to note that the calculation is performed using the hand position relative to the object, not relative to the camera.
Due to the unmeasurable body deformations of musculoskeletal humanoids, the position of the camera moves with the robot motion, and it is difficult to know its position accurately.
Therefore, it is effective to use the relative hand position of the object under the assumption that the position of the object to be manipulated is constant relative to the robot body.

\subsection{Self-body image acquisition procedure}
The muscular-articular relationship consists of a neural network with one intermediate layer that takes muscle length $l$ and muscle tension $t$ as input and joint angle $\theta$ as output \cite{kawaharazuka2020toward}.
The whole body of the musculoskeletal humanoid is classified into groups of functionally close muscles and joints in order to reduce the computational load for the purpose of ensuring real-time performance.
In this experiment, we consider the muscle-joint relationship for four groups: two groups of joints and muscles related to the scapula based on the trunk, and two groups of joints and muscles related to the shoulder and elbow based on the scapula.

%joint-muscle-mappingの初期モデルは、幾何モデルから生成される。
%幾何モデル上では、筋は予め設定する骨格に固定されたいくつかの経由点を直線で結ぶと仮定されている。
%従って、筋の伸び縮みや骨格の辺会は考慮されておらず、実機のjoint-muscle-mappingとは大きな誤差が生じる。
The initial model for joint-muscle-mapping is generated from the geometric model.
In the geometric model, the muscle is assumed to be connected by a straight line through some fixed points on the skeleton to be predetermined.
Therefore, it does not take into account the stretching and shrinking of the muscles and the edges of the skeleton, resulting in a large error compared to the actual joint-muscle-mapping.

\begin{figure}[tbh]
 \begin{center}
  \includegraphics[width=1.0\columnwidth]{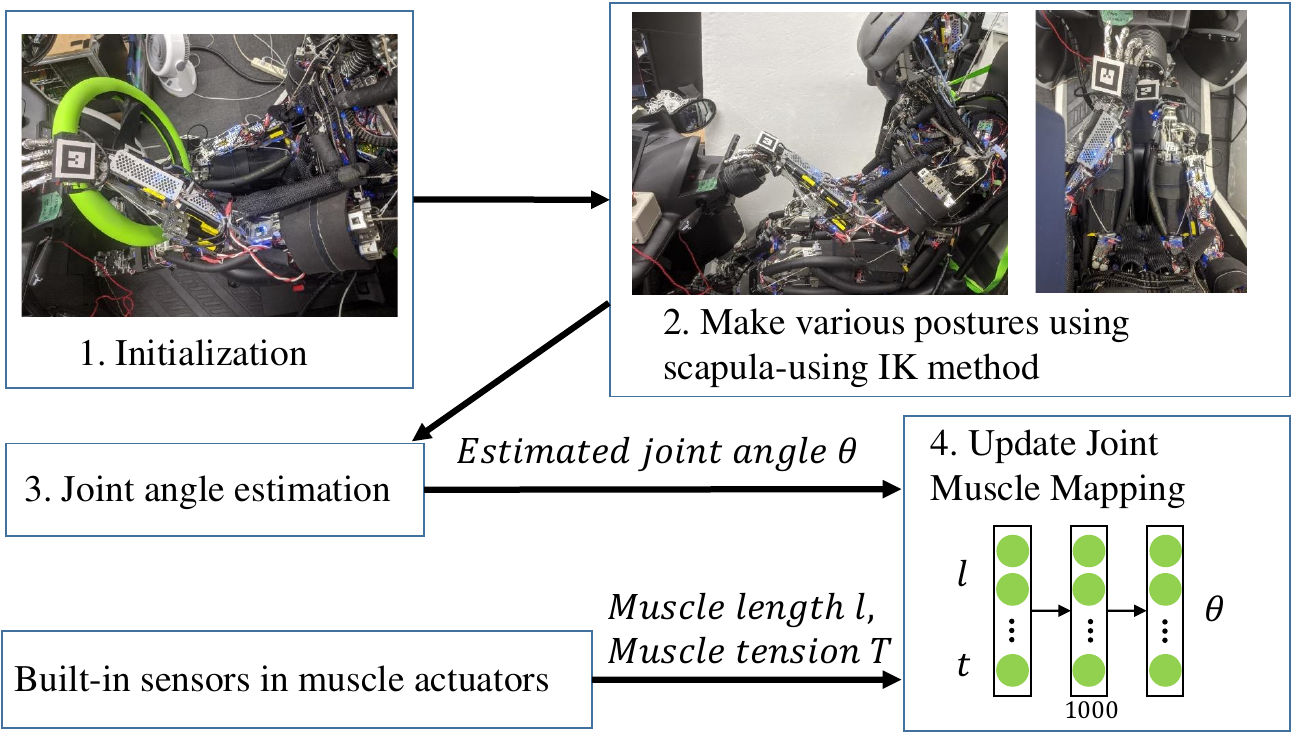}
  \caption{Joint Muscle Mapping learning method.}
  \label{figure:jmm_learning}
 \end{center}
\end{figure}

The outline of the learning method for the actual self-body image is shown in \figref{figure:jmm_learning}.
The concrete procedure is as follows.
\begin{itemize}
  \item initialize the posture from the previous section.
  \item Remove the steering wheel and attach the AR marker to the center of rotation of the steering wheel
  \item Apply the inverse kinematics method described in the previous section and have Kengoro assume various postures related to object manipulation in order to create teacher data.
  \item Using the joint angle estimation method described in the previous section, determine the hand position with respect to the trunk and the estimated joint angle.
  \item Accumulate a set of muscle length and muscle tension data acquired from the sensor in the muscle unit and the estimated joint angles as teacher data.
  \item Update the muscle-joint relationship using the teacher data\cite{kawaharazuka2019long}.
\end{itemize}

The key point here is that the same inverse kinematics method is used for the pose generation and joint angle estimation for the teacher data generation.
Due to the joint redundancy of the upper limbs of musculoskeletal humanoids, it is not possible to uniquely determine the estimated joint angle for a single hand position.
Therefore, we intend to absorb this discrepancy by using inverse kinematics that considers the same scapula in posture generation and joint angle estimation.
In the end, it is possible to acquire a self-body image that can move a hand to an appropriate position for an object.

\section{Steering Wheel Operation Experiment with Musculoskeletal Humanoid}
In order to verify the effectiveness of the aforementioned methods, we conducted an experiment using a full-body musculoskeletal humanoid, Kengoro, to operate a car at the wheel.
Humanoid driving can be used as is without any special modification to a car, and is useful in that it can perform multiple roles, such as carrying objects, in addition to driving, compared to normal automatic driving.
This is considered to be a useful feature of automated driving, as it can be used as-is without any special modifications to the car and can perform multiple roles such as transporting objects in addition to driving.
Among them, steering wheel operation is an excellent example of upper limb movements involving the shoulder complex, since it requires the simultaneous use of both arms and the application of appropriate force to the steering wheel.

\begin{figure}[tbh]
 \begin{center}
  \includegraphics[width=1.0\columnwidth]{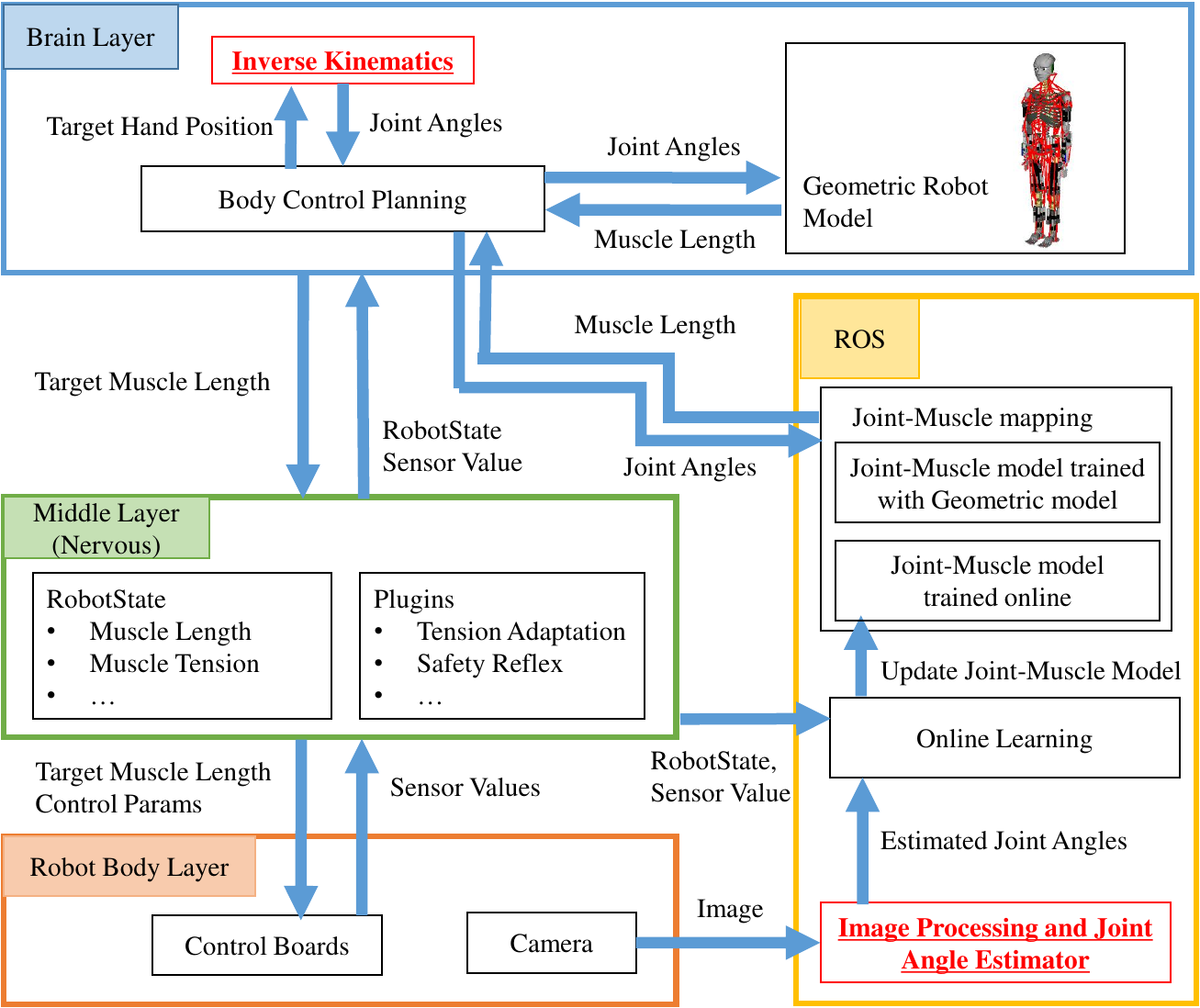}
  \caption{Whole software system including inversekinematics method and joint-angle estimation method.}
  \label{figure:whole_system}
 \end{center}
\end{figure}

The whole system in this experiment is shown in \figref{figure:whole_system}.
The Inverse Kinematics shown in red is the posture generation method in Section III, and the Image Processing and Joint Angle Estimatior is the robot posture estimation for self-body image learning in Section IV.
When learning the self-body image of the actual robot, $Online\ Learning$ is performed based on the estimated joint angle and the muscle-joint relationship is updated.

\begin{figure}[H]
 \begin{center}
  \includegraphics[width=1.0\columnwidth]{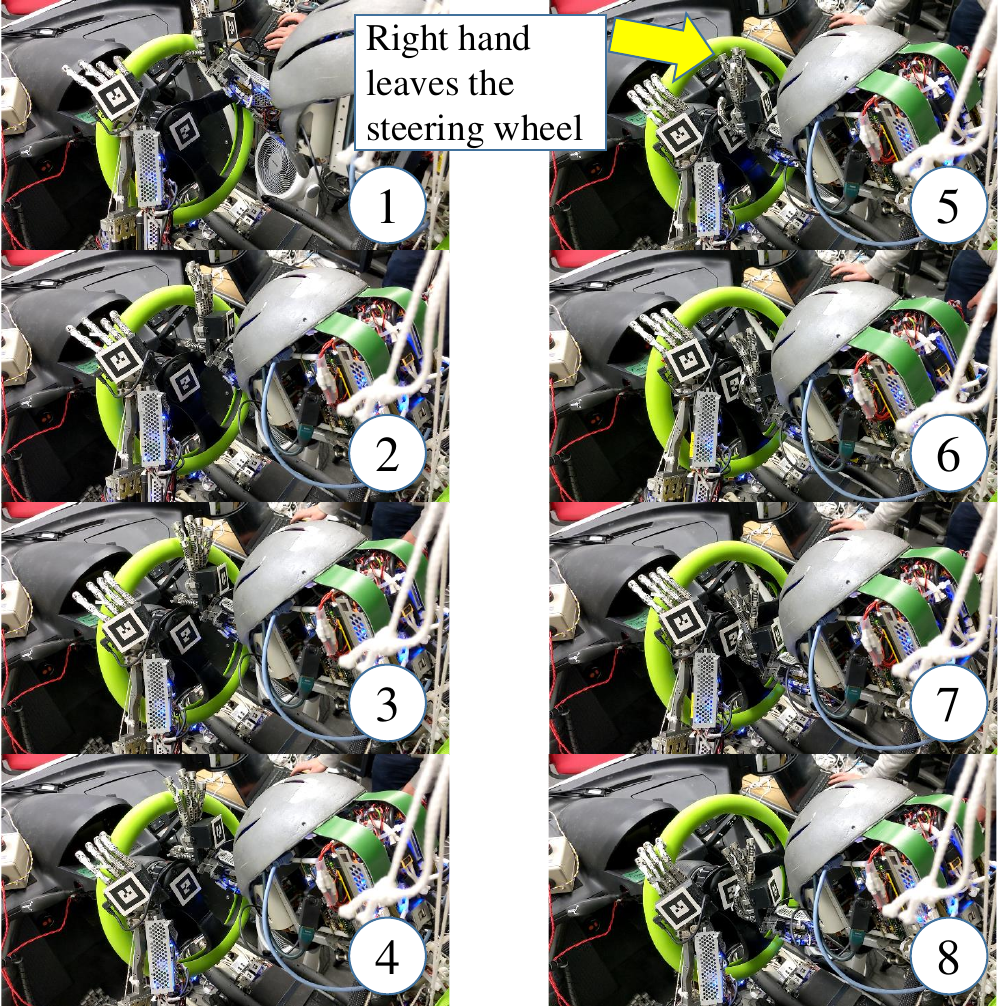}
  \caption{Handle operation experiment before online Joint-Muscle mapping learning.}
  \label{figure:handle_rotation_failure}
 \end{center}
\end{figure}

First, we show the steering wheel manipulation experiment in \figref{figure:handle_rotation_failure} without the self-body image acquisition described in Section IV.
It can be seen that the robot's posture deviates from the target posture due to the unmeasured deformation of the robot and high muscle redundancy, and that the hand comes off the handle immediately after the start of the experiment.

\begin{figure}[H]
 \begin{center}
  \includegraphics[width=1.0\columnwidth]{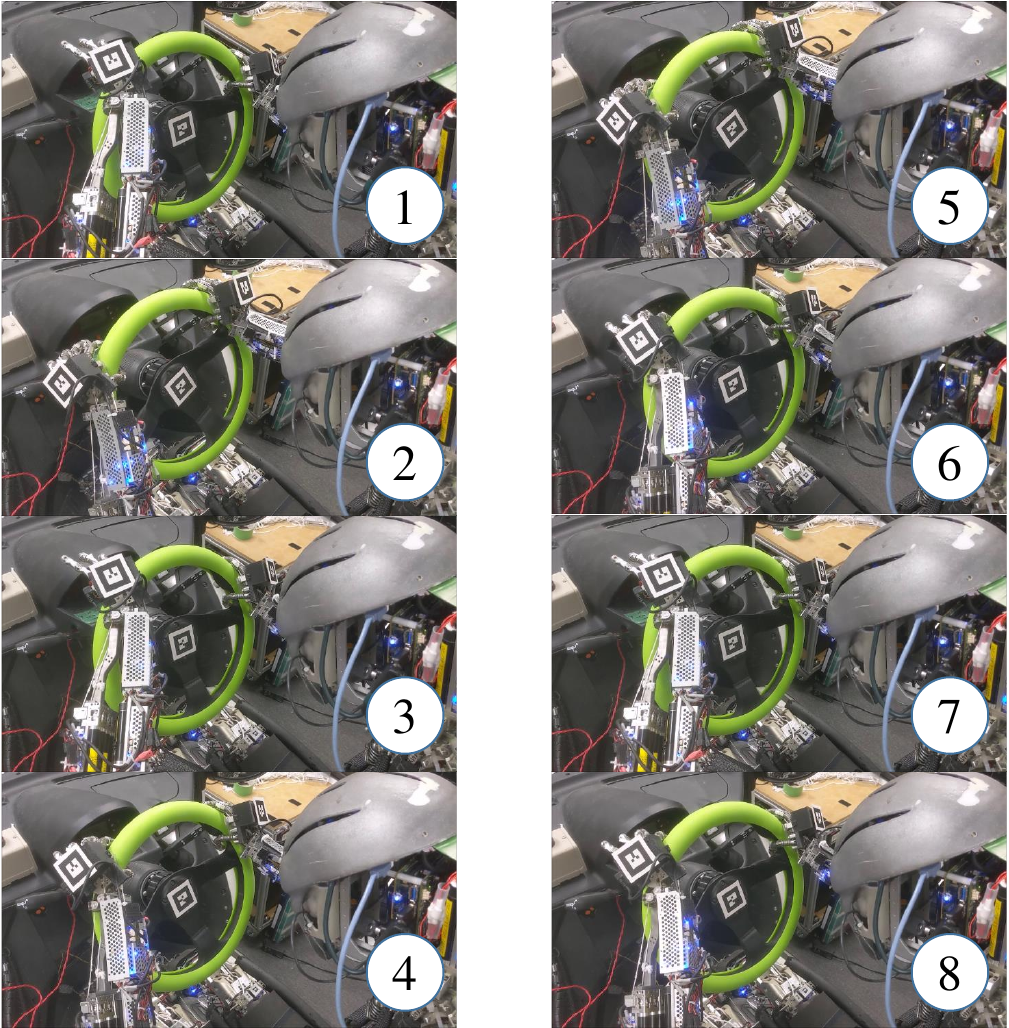}
  \caption{Handle operation experiment after online Joint-Muscle mapping learning.}
  \label{figure:handle_rotation_success}
 \end{center}
\end{figure}

\begin{figure}[H]
 \begin{center}
  \includegraphics[width=1.0\columnwidth]{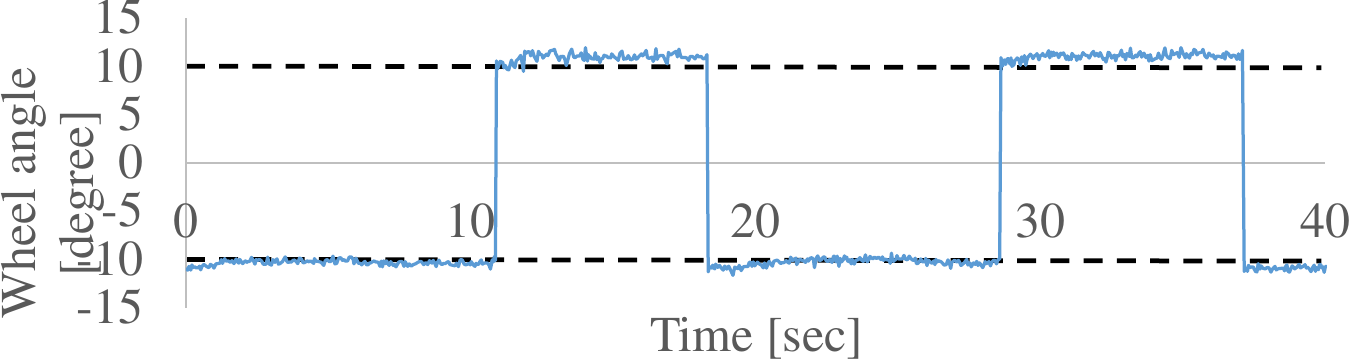}
  \caption{Time variation of handle rotation angle.}
  \label{figure:handle_rotation_angle}
 \end{center}
\end{figure}

Next, the experiment of steering wheel operation after learning the self-body image is shown in \figref{figure:handle_rotation_success}.
Since the robot posture is estimated based on the relative position of the steering wheel, we can see that the robot is able to generate muscle length commands to realize an appropriate posture for steering wheel manipulation.
The graph of the handle rotation angle obtained from the AR marker attached to the center of the handle is shown in \figref{figure:handle_rotation_angle}.
Both clockwise and counterclockwise rotations are stable at 10 degrees each.

\section{CONCLUSION}
The purpose of this study is to propose a hardware design and control strategy for task realization of a musculoskeletal humanoid.

For the hardware design, we aimed to increase the muscle redundancy and stabilize the joints by implementing the deep muscles, which have been neglected in the past due to spatial limitations.
Next, we proposed a posture generation method inspired by the scapulohumeral rhythm of the human body, taking advantage of the joint redundancy of the shoulder complex, which is driven by the scapula, and confirmed that this method can reduce the load compared to the scapulohumeral joint alone.
Then, we proposed a posture generation method based on an object manipulation.
Next, we proposed a method to absorb the difficult-to-measure deformations of musculoskeletal humanoids for the purpose of object manipulation, and to enable the learning of self-body images on actual machines.
Finally, we confirmed the effectiveness of these methods in a car steering wheel operation experiment by Kengoro.

The following three points can be considered as future tasks and prospects.
\begin{itemize}
  \item Coexistence of non-muscle planar structures and muscle actuators\\
    In musculoskeletal humanoids, it is considered that there is a limit to achieving the degree of freedom and stability of joints in the human body only by devising muscle arrangement.
    Therefore, it is considered that tissues other than muscles, such as ligaments and joint capsules, can exert appropriate force by elasticity and play a role in anti-gravity action and joint stabilization with less energy, and it is necessary to imitate these tissues in the human body.
  \item Development of self-body recognition method\\
    For the evaluation of human mimetic structures and the study of general control strategies, it is essential to develop a method for musculoskeletal humanoids to recognize their own states.
    It is considered to integrate the information from the external camera and the self-positioning sensor \cite{intel_realsense} in the robot.
  \item Application to general-purpose behavior\\
    In the proposed learning method of joint-muscle-mapping, the positional relationship between the robot and the manipulated object was assumed to be fixed and the initial posture was determined by a person.
    On the other hand, in many tasks, the object is not fixed and at the same time the robot itself needs to move.
    In order to apply our method to such general-purpose movements, a possible solution is for the robot to automatically find a reference from the surrounding environment.
    For example, when a robot is engaged in assembly work in a factory, it will handle the attached parts based on the assembly target.
\end{itemize}

%汎用的な動作への適用
%提案するjoint-muscle-mappingの学習手法では、ロボットと操作物体の位置関係が固定され、初期姿勢を人が決めることが前提となっていた。
%一方、多くのタスクは物体が固定されていないと同時に、ロボット自体が移動する必要がある。
%そのような汎用動作に本手法を適用するために、ロボットが周囲の環境から自動的に基準を見つけ出すといった解決方法が考えられる。
%例えばロボットが工場の組み立て作業に従事する場合、組付け対象を基準とし取り付け部品をハンドリングすることになる。

\addtolength{\textheight}{-12cm}   % This command serves to balance the column lengths

%% \input src/appendix.tex
%% \input src/acknowledge.tex

\bibliographystyle{junsrt}
\bibliography{main}

\end{document}